\pdfoutput=1
\documentclass{article}
\usepackage[numbers]{natbib}
\usepackage[preprint]{neurips_2026}
\makeatletter
\renewcommand{\@noticestring}{Preprint.}
\makeatother
\usepackage[utf8]{inputenc}
\usepackage[T1]{fontenc}
\usepackage{float}
\usepackage{hyperref}
\usepackage{url}
\usepackage{booktabs}
\usepackage{array}
\usepackage{amsfonts}
\usepackage{amsmath}
\usepackage{graphicx}
\usepackage{microtype}
\title{Three Ways Classical Test Theory Can Mislead About LLM Judges}

\author{%
  Louis Yiven Zhu\\
  University of Oxford\\
  \texttt{yiven.zhu@oii.ox.ac.uk}
}

\begin{document}
\maketitle

\begin{abstract}
Evaluations that use a large language model (LLM) as a judge have begun to borrow
reliability statistics from classical test theory and its extensions. We examine three
such statistics that need one administration and no gold labels. None of them can isolate
the judge, because one judge under one prompt supplies no variance component of its own.
Claude Haiku 4.5 judged 210 constructed short answers against ten-element checklists. On
the 180 with parsed verdicts, the Kuder--Richardson coefficient (KR-20) came out at
$0.5223$ on the judge's verdicts and $0.5231$ on error-free gold verdicts. In simulation,
bank design alone moves KR-20 from $0.01$ to $0.68$ at the judge's measured $4.72\%$
error rate. The dependability index $\Phi(\lambda)$, a ratio of mean squared distances
from the pass mark, sits $0.22$ to $0.38$ below the judge's accuracy against gold and
returns $0.54$ to $0.68$ on error-free gold verdicts.
Livingston--Lewis accuracy treats the rubric elements as a sample, and at a pass mark of
five elements it credits error-free gold scores with $0.78$, close to the judge's $0.81$.
A statement about the judge therefore needs gold labels or a varied scorer facet, and a
reliability ratio needs the bank's spread beside it. One of the four closest
judge-evaluation papers varies the prompt and still reads a reliability below $0.7$ as a
sign that a model cannot serve as a judge, although that reliability moves with the
spread of the samples scored. We derive a decision table and four reporting lines from
these two rules.
\end{abstract}

\section{Introduction}

Reliability statistics borrowed from classical test theory can describe the bank of
responses an LLM judge scores while appearing to describe the judge. On the bank we
study, the Kuder--Richardson internal-consistency coefficient for binary scores
(KR-20~\cite{kuder1937}) comes out at $0.5223$ on the verdicts of Claude Haiku 4.5 and at
$0.5231$ on gold verdicts that contain no scoring error. A reader who took $0.52$ as the
judge's reliability would be reading a value that the bank largely sets, as an
error-free scorer returns the same value. The statistics themselves are sound, and our argument concerns
only the reading that attributes their values to the judge.

Evaluators now use one language model to score another's output across open-ended
tasks~\cite{zheng2023}. Documented failure modes, from position bias~\cite{wang2024} to a
preference for a model's own generations~\cite{panickssery2024}, give them good reason to
borrow tools built for measuring people. Recent work imports internal-consistency
coefficients and factor structure~\cite{feuer2025}, item response theory~\cite{choi2026}
and datasheet-style protocols~\cite{usami2026}. We read classical test theory broadly, to
include its generalisability and strong true-score extensions, as two of the three
statistics we examine come from those extensions.

\begin{figure}[t]
\centering
\includegraphics[width=\textwidth]{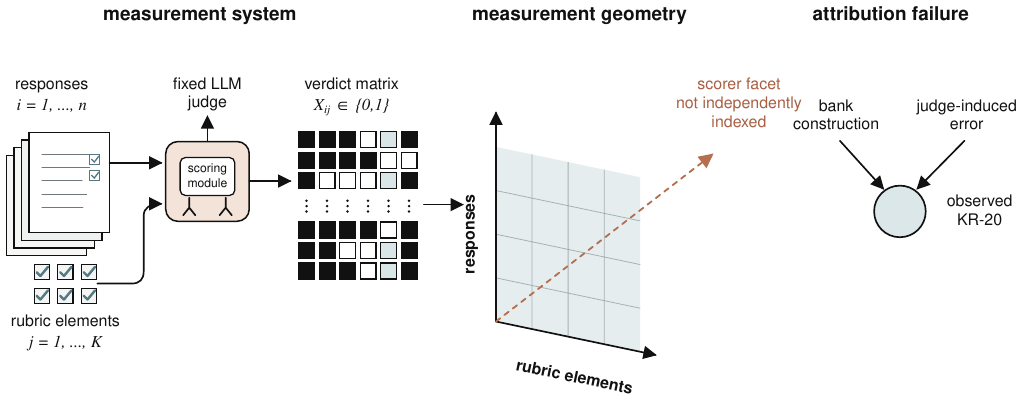}
\caption{Why KR-20 cannot identify the judge's reliability. A fixed LLM judge scores $n$
responses against $K$ binary rubric elements and fills the verdict matrix
$X_{ij}\in\{0,1\}$. KR-20 partitions the matrix's variance into components for responses,
elements and a residual. One judge under one prompt phrasing supplies no variance along a
scorer axis. Errors that the judge repeats within a response merge with the response
component, errors that it repeats across responses on one element merge with the element
component, and the rest merge with the residual. Bank construction and judge error can
thus produce the same coefficient, one that characterises the measurement system as a
whole.}
\label{fig:facets}
\end{figure}

Each borrowed statistic answers a question fixed by a measurement design. The design
specifies the sources of variance, called facets, that the data let vary. In the judge
setting the response takes the examinee's place, the rubric takes the test's and the judge
takes the scorer's (Figure~\ref{fig:facets}). In the designs behind the three statistics
we examine, the response is the object of measurement and one judge under one prompt is a
fixed, hidden condition of every verdict. Their data therefore hold no variance component
that belongs to the judge alone~\cite{brennan2001}. The judge's errors merge with the
components for responses, for rubric elements and for their interaction, and no statistic
computed from those components can separate them again. A question about the judge thus
concerns the one condition that these designs hold fixed.

We ask three questions about statistics that a practitioner might use at a deployment
gate, where a judge's score decides whether a response passes. \textbf{Q1} asks whether
KR-20 computed on a judge's verdicts measures the judge. \textbf{Q2} asks whether the
dependability index $\Phi(\lambda)$ of generalisability theory estimates the probability
that a gate decision is correct. \textbf{Q3} asks whether Livingston--Lewis
classification accuracy estimates the judge's agreement with gold. Each of the three needs a single administration and no gold labels. That property makes the statistics appealing, and it also prevents each from isolating the judge. We answer all three
negatively, and for different reasons. KR-20 moves with bank design, judge error and the
structure of that error, and one value cannot separate them (Section~\ref{sec:misuse1}).
$\Phi(\lambda)$ is a ratio of squared distances from the pass mark and returns $0.54$ to
$0.68$ for a scorer that never errs (Section~\ref{sec:misuse2}). Livingston--Lewis counts
the sampling of rubric elements as error and credits the same scorer with $0.75$ to $0.85$
(Section~\ref{sec:misuse2}).

We make four contributions to how teams report judge reliability. First, we compute each
statistic on the gold verdicts beside its value on the judge's verdicts, a check we call
the gold-computed null. The null is the value an error-free scorer returns on the same
bank. On our bank it matches the judge's KR-20 to two decimals and gives the error-free
scorer $\Phi(\lambda)$ values of $0.54$ to $0.68$ and Livingston--Lewis accuracies of
$0.75$ to $0.85$. Second, we separate three inputs that move KR-20, namely bank
design, the judge's error rate and the clustering of its errors. A joint simulation shows
that bank design dominates at realistic error rates, and the real bank shows that
clustered errors can offset the loss that independent errors would cause. Third, we show
that $\Phi(\lambda)$ and Livingston--Lewis accuracy each estimate something other than the
accuracy of a gate decision, and that element sampling dominates the second mismatch on
our bank. Fourth, we give a decision table that matches practitioner questions
to the quantities that answer them, four reporting lines, and a phrasing design that
identifies the scorer facet even for a deterministic judge.

Related work has shown particular agreement measures failing particular uses.
Landesberg~\cite{landesberg2026} shows that global judge--human agreement can mispredict
the quality of best-of-N selection, and Chang et al.~\cite{chang2026} show that an
aggregate agreement gate can pass while per-dimension claims fail. Choi
et al.~\cite{choi2026} measure a judge's stability across prompt variants, and Schroeder
and Wood-Doughty~\cite{schroeder2024} measure it across repeated samples. We ask a prior
question, namely which facet each borrowed reliability statistic partitions and which true
score it refers to. The mechanisms behind our answers are standard
psychometrics~\cite{gulliksen1950,brennan2001,livingston1995}. Our contribution is to show
where each one enters judge evaluation and to give a check, the gold-computed null, that
any team with labels can run. Section~\ref{sec:setup}
describes the bank and the judge, Section~\ref{sec:misuse1} answers Q1, and
Section~\ref{sec:misuse2} answers Q2 and Q3. Section~\ref{sec:gating} reports what the
four nearest papers do and turns the answers into a decision table and reporting lines,
Section~\ref{sec:limits} states the limitations, and Section~\ref{sec:conclusion}
concludes.

\section{Setup}
\label{sec:setup}

We built the bank to supply a gold verdict for every element without annotation error,
and to keep surface cues weak. It holds 210 constructed short answers to 15 science
questions, each paired with a checklist of $K=10$ elements written for its question. Each
element states one checkable claim and has a purpose-written distractor, a sentence that
asserts something adjacent on the same sub-topic. A response includes a random subset of
the elements, and each absent element's slot takes its distractor with probability $0.65$
and otherwise stays empty. Length still carries information, and sentence count
correlates with the gold total at $r=0.63$. Within a slot, however, a distractor resembles
its element in length and topic and differs in content. Gold totals have mean $5.45$ and
SD $2.21$ on the $0$--$10$ scale, and 6 of the 210 responses reach the ceiling of 10. We
made the bank harder than an earlier version, on which judges from the same model family
reached correlations of $0.99$ to $1.00$ with gold and left no error to study.
Appendix~\ref{app:bank} shows one response in full with its gold and judge verdicts.

Claude Haiku 4.5 (\texttt{claude-haiku-4-5-20251001}) served as the judge at temperature
$1.0$, the API default, and returned a YES or NO verdict for each element. The judge is an instrument in
this design, and one model suffices for the argument, as the analytic claims do not
depend on which model fills the role (Section~\ref{sec:limits}). Its output parsed into
ten clean verdicts for 180 of the 210 responses. The 30 failures fall in three questions,
and two of those questions lost all 14 of their responses. Because responses to one
question share an element set, every interval we report resamples the 13 question
clusters that remain. On the 180 scored responses, judge and gold totals correlate at
$r=0.921$ and the judge errs on $4.72\%$ of element verdicts ($95\%$ cluster interval
$2.6\%$--$6.9\%$). The judge is lenient by $0.46$ elements on average. It accepts an
absent element $10.4\%$ of the time ($5.5\%$--$15.5\%$) and rejects a present one $0.1\%$
of the time ($0.0\%$--$0.3\%$). The bank thus leaves the judge enough error to study, and
that error runs toward acceptance and clusters by question (Appendix~\ref{app:bank}).

\section{Internal consistency has no scorer facet}
\label{sec:misuse1}

We answer Q1 negatively, because KR-20 cannot separate the judge's accuracy from the
design of the bank it scores. The coefficient needs one administration and no labelled
subset, and any team with a rubric-scoring judge can compute it from verdicts it already
holds.
Computing it on the same responses from the judge's verdicts $X$ and from the gold
verdicts $G$ gives
\[
\mathrm{KR}\text{-}20(X) = 0.5223, \qquad \mathrm{KR}\text{-}20(G) = 0.5231 .
\]
A cluster bootstrap over the 13 question clusters puts the $95\%$ interval for the
difference at $[-0.073,\,+0.074]$. We do not read the result as showing that judge error
leaves KR-20 unchanged, as thirteen clusters cannot exclude a difference of $0.05$.
The comparison establishes a narrower fact, namely that an error-free scorer and a judge
that errs on $4.72\%$ of verdicts produce coefficients within $0.001$ of each other on
this bank.

The result follows from the measurement design, since KR-20 partitions variance across
responses and rubric elements and gives the judge no term of its own. Judge error can
therefore move the coefficient in either direction. Errors confined to one response and
element lower it, whereas errors that a judge repeats within a response merge with the
response component and can raise it. Applying independent errors at the judge's measured
rates to the gold verdicts gives a mean coefficient of $0.470$ (SD $0.026$), and the
observed $0.5223$ exceeds all but $1.7\%$ of such draws. The allocation of this judge's
errors, which cluster by question and within responses, lifts the coefficient from
$0.470$ to $0.522$ (Appendix~\ref{app:bank}). The near-equality with gold is thus partly a
coincidence of error structure. The gold value follows from the bank's construction
alone. The standardised form $K\bar\rho/(1+(K-1)\bar\rho)$ returns $0.520$ at the gold
verdicts' mean inter-element correlation of $\bar\rho = 0.098$, and the judge's errors
leave that correlation unchanged to three decimals.
Attributing any part of the value to the scorer would require a rater
facet~\cite{brennan2001} or a labelled subset.

\begin{figure}[t]
\centering
\includegraphics[width=0.88\textwidth]{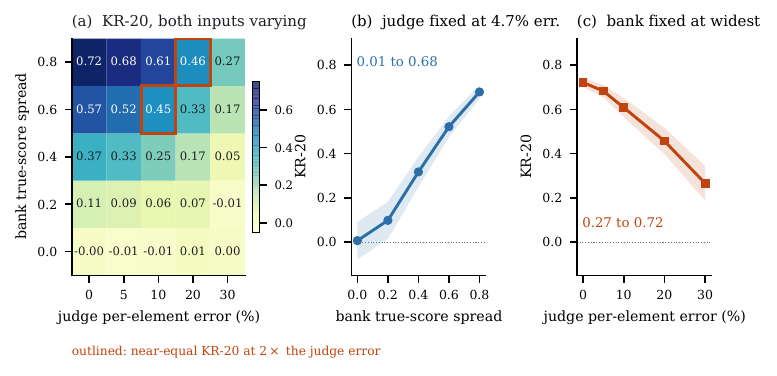}
\caption{KR-20 responds to bank design and to judge error, and bank design dominates at
realistic error rates. \textbf{(a)} Full grid, 60 replicates per cell; outlined cells
have near-identical coefficients ($0.451$ and $0.457$) at twice the judge error.
\textbf{(b)} At the measured $4.72\%$ error rate, bank design alone moves the coefficient
across most of its range ($\pm 1$ SD shaded). \textbf{(c)} At the widest bank, judge error
lowers the coefficient by $0.04$ at $5\%$ and by $0.46$ at $30\%$.}
\label{fig:sweep}
\end{figure}

Bank design alone can move KR-20 across most of its range while judge error stays fixed.
We varied the bank's spread from $0.0$ to $0.8$ and judge error from $0\%$ to $30\%$, with
60 replicates per cell (Figure~\ref{fig:sweep}). Spread is the width of the uniform range
from which we draw each response's element-presence rate, and it controls
between-response true-score variance while holding the mean fixed
(Appendix~\ref{app:grids}). At the measured error rate of $4.72\%$, mean KR-20 rises from
$0.007$ at spread $0.0$ through $0.098$, $0.317$ and $0.522$ to $0.679$ at spread $0.8$
(Table~\ref{tab:sweep}). The same judge at the same error rate can return any value from $0.01$ to $0.68$, depending on how we build the bank. The mechanism is range
restriction, familiar since Gulliksen~\cite{gulliksen1950}, whereby a bank with less
true-score variance yields a lower coefficient at any level of scoring accuracy.

Judge error moves the coefficient far less than bank design at realistic rates, and the
coefficient still cannot separate the two effects. At the widest bank, raising judge error
from $0\%$ to $5\%$ lowers KR-20 by $0.04$ and raising it to $30\%$ lowers it by $0.46$.
Bank construction moves it by $0.72$ at zero error. Doubling the judge's error can also
leave the coefficient nearly unchanged if the bank widens. A spread of $0.6$ at $10\%$
error gives $0.451$, and a spread of $0.8$ at $20\%$ gives $0.457$
(Appendix~\ref{app:grids}). The simulated
judge errs independently and symmetrically, and replacing its flips with one-directional
errors at the measured rates leaves the range at $0.01$ to $0.67$. The real judge's
clustering adds a third input that the value cannot separate from the other two. A
reliability value computed over rubric elements thus describes the bank and the judge
jointly, and nothing in the value apportions it between them.

\section{Two estimand mismatches}
\label{sec:misuse2}

The answers to Q2 and Q3 are also negative, as two further statistics look like the accuracy of a gate decision and measure something else. The dependability index $\Phi(\lambda)$ of
generalisability theory~\cite{brennan2001} takes a pass mark $\lambda$ and lies in
$[0,1]$. We estimate it by Brennan's procedure for a design that crosses responses with
element positions (Appendix~\ref{app:estimators}), with $\lambda=c/K$ the proportion of the $K$
elements that a response must satisfy to pass. Because the probability of a correct gate
decision shares both properties, readers can take the index for that probability. It is
a ratio of mean squared distances from the pass mark, however, whereas classification
accuracy is a probability. At pass marks of $c=4$ to 7 elements the judge's accuracy against gold is
$0.961$, $0.928$, $0.900$ and $0.917$, while $\Phi(\lambda)$ is $0.741$, $0.600$, $0.516$
and $0.608$ (Figure~\ref{fig:estimands}a, with accuracy intervals in Table~\ref{tab:ll}).

Two checks trace the resulting gaps of $0.22$ to $0.38$ to the estimand. On the error-free
gold verdicts, whose accuracy is 1 by construction, $\Phi(\lambda)$ is $0.676$, $0.543$,
$0.541$ and $0.673$. An index that returns $0.54$ to $0.68$ for a scorer that never errs
cannot estimate the probability that a gate decision matches gold. In data simulated from a
beta-binomial model with the bank's fitted parameters, an accuracy estimate built on that
model recovers the true-score accuracy to within $0.02$ at every cut, whereas
$\Phi(\lambda)$ departs from it by $0.14$ to $0.25$ (Appendix~\ref{app:grids}). The real
bank shows a departure of the same size, as $\Phi(\lambda)$ sits $0.14$ to $0.24$ below
Livingston--Lewis accuracy. The remaining $0.08$ to $0.17$ of each gap separates
Livingston--Lewis accuracy from accuracy against gold and belongs to the second
mismatch.

Livingston--Lewis accuracy~\cite{livingston1995} carries a second estimand mismatch, and
on our bank element sampling dominates it. We implement the procedure in its
two-parameter form, a beta-binomial model with true scores on $[0,1]$ and a test length of
$K$. The fitted model passes a goodness-of-fit test on the judge's totals
(Appendix~\ref{app:bank}). It estimates how often a response's observed pass or fail
matches the classification of its own true score. That true score is the proportion of a
wider domain of elements that the response satisfies, and the $K$ elements in the rubric
are a sample from that domain. The estimand suits a testing programme that treats its
items as one draw from a domain and asks whether a pass would survive another draw, and
Table~\ref{tab:decisions} keeps it for that question. A judge auditor asks whether the
judge's decision matches gold on the elements actually used.

The element-sampling error that the model finds in the error-free gold totals is a
property of the bank, and on our bank it exceeds the whole difference between the two
accuracies. The beta-binomial fitted to the
judge's totals implies a reliability of $0.523$, close to KR-20 on either set of verdicts.
Its error term reflects the same element sampling that KR-20 measures. Applied to the
error-free gold totals, Livingston--Lewis returns accuracies of $0.846$, $0.781$, $0.750$
and $0.769$ at cuts 4 to~7 (Table~\ref{tab:ll}). An error-free scorer thus falls $0.15$ to
$0.25$ short of perfect accuracy on this estimand. The judge's own estimates of $0.882$,
$0.812$, $0.759$ and $0.750$ lie within $0.04$ of the error-free scorer's at every cut. With
$A_J$ the judge's accuracy against gold and $\mathrm{LL}_J$ and $\mathrm{LL}_G$ the two
Livingston--Lewis estimates, the difference of $0.08$ to $0.17$ decomposes exactly as
\[
A_J-\mathrm{LL}_J=(1-\mathrm{LL}_G)-(1-A_J)-(\mathrm{LL}_J-\mathrm{LL}_G).
\]
The error-free shortfall $1-\mathrm{LL}_G$ runs from $0.15$ to $0.25$, the judge's gate
error $1-A_J$ from $0.04$ to $0.10$, and $\mathrm{LL}_J-\mathrm{LL}_G$ from $-0.02$ to
$0.04$. Element sampling contributes more than the whole difference, and the judge's own
errors offset part of it.

A validity gap, meaning a judge true score that departs from gold, pushes the same
comparison the other way. In a simulation with no element sampling, where the judge's true
score departs from gold, accuracy against that true score exceeds accuracy against gold by
$0.033$ to $0.080$ across cuts 4 to~7 (Figure~\ref{fig:estimands}b). A difference between
Livingston--Lewis accuracy and accuracy against gold therefore mixes the bank's element
sampling with the judge's errors and any validity gap. The same estimate computed on gold
approximates the element-sampling part on its own. The wider family of single-administration
indices~\cite{subkoviak1976,hanson1990,lee2002,peng1980} estimates decision consistency
across parallel administrations or accuracy against a model true score, and none refers
to gold on the elements used. An auditor needs the
probability that the judge's gate decision matches gold, and neither $\Phi(\lambda)$ nor
Livingston--Lewis accuracy estimates it.

\section{Current practice and what to report}
\label{sec:gating}

One of the four judge-evaluation papers closest to ours already reads a reliability level
as judge capability. Feuer et al.~\cite{feuer2025} compute Cronbach's $\alpha$
per rubric factor with questions as the items and compare it across four judges on one
benchmark. Holding the benchmark fixed removes the bank from that comparison, although
$\alpha$ need not rank judges by accuracy, as clustered errors can raise it
(Section~\ref{sec:misuse1}). Usami et al.~\cite{usami2026} and Schwinn
et al.~\cite{schwinn2026} mention neither internal consistency nor Cronbach's~$\alpha$.
Choi et al.~\cite{choi2026} come closest to measuring the scorer facet, since their graded
response model gives each prompt variant its own item parameters. They also read a marginal reliability
below $0.7$ as a sign that a model ``lacks fundamental capability to serve as a judge''.
That reliability divides the variance of latent quality across the samples scored by
itself plus estimation error, and its level inherits the range restriction of
Section~\ref{sec:misuse1}. Varying the scorer facet identifies the judge's contribution
and leaves the bank's influence on any ratio built from it. None of the four commits the three errors in the form we define,
and four papers cannot support a claim about the field. We accordingly claim only that
borrowed reliability statistics are arriving in judge evaluation, and that one published
reading already credits the judge with a level that the scored samples help set.

Table~\ref{tab:decisions} turns the answers to Q1 to Q3 into decisions. Each row pairs a
question that a practitioner asks with the quantity that answers it, what that quantity
requires, and the tempting substitute that answers a different question. Every question
about the judge itself requires gold labels or varied phrasings. The label-free
statistics answer questions about the bank and the checklist, and a practitioner without
labels or a phrasing study can report them only in those terms.

\begin{table}[t]
\centering
{\small
\begin{tabular}{>{\raggedright\arraybackslash}p{3.7cm}>{\raggedright\arraybackslash}p{3.3cm}>{\raggedright\arraybackslash}p{1.45cm}>{\raggedright\arraybackslash}p{3.65cm}}
\toprule
Question & Quantity that answers it & Requires & Tempting substitute \\
\midrule
How often does the judge's pass or fail match gold? & Accuracy against gold, with
cluster intervals & Gold labels & $\Phi(\lambda)$, a ratio of squared distances, or
Livingston--Lewis accuracy, which counts element sampling as error \\
\addlinespace
Where does the judge err, and in which direction? & False-accept and false-reject rates
per checklist line & Gold labels & A whole-rubric reliability coefficient \\
\addlinespace
Do the judge's decisions depend on the wording of its instruction? & Phrasing variance
components from a person-by-phrasing study (Appendix~\ref{app:phrasing}) & Varied
phrasings & Agreement
across repeated samples at one prompt, which measures decode-time noise \\
\addlinespace
Do the checklist elements rank responses consistently? & KR-20 & Neither & KR-20 on
judge verdicts read as a property of the judge \\
\addlinespace
Would pass or fail survive another sample of checklist elements? & Livingston--Lewis
consistency and accuracy & Neither & Accuracy against gold, which holds the elements
fixed \\
\bottomrule
\end{tabular}}
\caption{Questions that a practitioner asks, and the quantities that answer them. The
first three questions concern the judge and require gold labels or varied phrasings. The
last two concern the bank and the checklist, and the label-free statistics answer them.}
\label{tab:decisions}
\end{table}

Four reporting lines carry Table~\ref{tab:decisions} into any disclosure that accompanies
a deployed judge. The first gives the bank's characteristics wherever a reliability value
appears, meaning $K$, the numbers of responses and question clusters, the SD of gold
totals and the mean inter-element correlation. The second places every coefficient beside
its value on gold, at the cost of one line of code. The third names $\Phi(\lambda)$ a
dependability coefficient and keeps it clear of numerical comparison with any accuracy.
The fourth reports the judge's false-accept and false-reject rates against gold, overall
and per checklist line, with intervals clustered on question.

Only the fourth line describes the judge directly. On this bank it reports a judge that
accepts $10.4\%$ of absent elements and rejects $0.1\%$ of present ones, and five of its
130 question-specific checklist lines hold 25 of its 85 errors. A team reading those rates
learns which lines to rewrite, and that a gate on judge scores passes more responses than
the same gate on gold. A team with gold labels on a subset should report the second and
fourth lines on that subset and give the first line for the full bank from the judge's
verdicts, saying so. Appendix~\ref{app:template} gives worked wording with this study's
values.

\section{Limitations}
\label{sec:limits}

The largest limitation is that we name the scorer facet as the missing quantity and do not
measure it. Measuring it requires a generalisability study that holds the checklist and
responses fixed and paraphrases the scoring instruction, since a deployment commits to one
phrasing and inherits its variance. Choi et al.~\cite{choi2026} measure a related facet
through prompt variants. Appendix~\ref{app:phrasing} specifies a design that estimates the
phrasing component and reports $\Phi$ at a single phrasing, and it shows that three
phrasings leave the component imprecise. Repeated administration at a fixed prompt would not suffice,
because it measures decode-time sampling noise and holds the phrasing fixed. We also scored
each response once at temperature $1.0$. Without a second run we cannot say how often one
would change a verdict on this bank, although an earlier 20-response pilot with $K=8$
returned identical scores across two administrations.

The remaining limitations narrow the illustrations and leave the analytic claims intact.
Our results rest on one judge from one model family and say nothing about error that
correlates within a family. With canonical element sentences repeated verbatim across the
responses to a question, the task sits closer to sentence detection than to factual
adjudication. The $4.72\%$ error rate plausibly understates what a realistic rubric would
produce. Parse failures remove 30 of the 210 responses, including every response to two
questions, and those questions may not be missing at random. Counting every unparsed
verdict as an error raises the per-element error rate to an upper bound of $18.3\%$. At a
pass mark of five elements, treating an unparsed output as a rejection gives an accuracy
of $0.852$ over all 210 responses and counting it as a gate error gives $0.795$, against
$0.928$ on the scored responses (Appendix~\ref{app:repro}). The effective sample is 13
clusters, as responses to one question share an element set, and every interval we report
is correspondingly wide. We did not pre-register these analyses, and the released code
recomputes every quantity measured on the bank exactly while reproducing the simulation
grids statistically (Appendix~\ref{app:repro}).

\section{Conclusion}
\label{sec:conclusion}

Statistics computed from one judge's verdicts under one prompt describe the bank as
readily as the judge. On our bank KR-20 matches its error-free value to two decimals,
$\Phi(\lambda)$ sits $0.22$ to $0.38$ below the accuracy it resembles, and
Livingston--Lewis credits an error-free scorer with only $0.75$ to $0.85$. Measurement
theory locates validity in the interpretation of a score~\cite{messick1990,kane1992}, and
each misreading attaches to the judge a value that the bank helps set. The measurement
design fixes the indexing of a statistic, meaning which facet it partitions and which true
score it refers to, before anyone chooses an estimator. No care in estimation recovers the
indexing afterwards. A statement about the judge therefore needs a quantity that belongs to
the judge, and only gold labels or a varied scorer facet can supply one. A reliability ratio
built from such a quantity still needs the bank's spread beside it. Teams with labels can
compute each coefficient on gold beside the judge's value and report the judge's error
rates per checklist line. Teams without them can run the phrasing study of
Appendix~\ref{app:phrasing}, report its phrasing component and describe every coefficient
as a property of the whole measurement system. Appendix~\ref{app:repro} links the bank, estimators,
simulations and regeneration scripts.

\clearpage
\appendix
\setcounter{figure}{0}\renewcommand{\thefigure}{A.\arabic{figure}}
\setcounter{table}{0}\renewcommand{\thetable}{A.\arabic{table}}
\renewcommand{\theHfigure}{A.\arabic{figure}}\renewcommand{\theHtable}{A.\arabic{table}}
\section{Supplementary material}
\label{app:main}

Every number in the main text traces to a script or a table in this appendix.
Appendix~\ref{app:repro} describes the released artefacts and the exclusions, and
Appendix~\ref{app:bank} documents the bank and the judge's errors behind
Section~\ref{sec:setup}. Appendix~\ref{app:estimators} states the estimators, and
Appendix~\ref{app:grids} gives the simulation grids and controls behind
Sections~\ref{sec:misuse1} and~\ref{sec:misuse2}. Appendix~\ref{app:phrasing} specifies
the scorer-facet study that Section~\ref{sec:limits} calls for,
Appendix~\ref{app:template} gives worked wording for the reporting lines of
Section~\ref{sec:gating}, and Appendix~\ref{app:glossary} defines the terms and notation
(Table~\ref{tab:glossary}).

\subsection{Released artefacts and reproduction}
\label{app:repro}

The repository at \url{https://github.com/louisyzhu/llm-judge-reliability} holds the bank,
the estimators, the simulations and the scripts that regenerate every number and figure
(Table~\ref{tab:artefacts}). Reproducibility differs by quantity, and the script
\texttt{bank\_stats.py} recomputes every quantity measured on the real bank exactly from
the raw verdicts. The simulation grids reproduce statistically, as we generated the published grids on a random stream that differs from the one the released script draws.
Re-simulating moves individual cells by at most $0.31$ of a per-cell standard deviation,
and the figures regenerate from the published values in \texttt{sweep\_grid.json}. Judge
scoring used \texttt{claude-haiku-4-5-20251001} at temperature $1.0$, and
Appendix~\ref{app:estimators} lists every seed.

\begin{table}[H]
\centering
{\small
\begin{tabular}{llp{6.6cm}}
\toprule
Artefact & File & Contents \\
\midrule
Bank & \texttt{judge\_item\_bank.csv} & 210 responses with question id, question and answer
text, element text, element-level gold ($G_{ij}$) and judge ($X_{ij}$) verdicts and
totals; judge fields blank for the 30 unparsed responses \\
Estimators & \texttt{estimators.py} & KR-20, KR-21, inter-element correlation,
beta-binomial MLE and goodness of fit, Livingston--Lewis DC and CA, variance components
and $\Phi(\lambda)$, cluster bootstrap \\
Bank statistics & \texttt{bank\_stats.py} & Every quantity measured on the real bank,
with cluster intervals and the independence check of Section~\ref{sec:misuse1} \\
Simulations & \texttt{sweeps.py} & Two-way sweep, measured-rate run, one-directional
sweep, $\Phi(\lambda)$ control, Livingston--Lewis estimand control, phrasing
identification check \\
Published grids & \texttt{sweep\_grid.json} & Grid values as reported here, including the
60-replicate run at the measured $4.72\%$ error rate \\
Figures & \texttt{make\_figures.py} & Regenerates Figures~\ref{fig:sweep},
\ref{fig:bank} and~\ref{fig:estimands} from the released data; Figure~\ref{fig:facets} is
a schematic \\
\bottomrule
\end{tabular}}
\caption{Contents of the released bundle. Running \texttt{python bank\_stats.py},
\texttt{python sweeps.py} and \texttt{python make\_figures.py} regenerates every quantity
and data figure with \texttt{numpy}, \texttt{scipy}, \texttt{pandas} and
\texttt{matplotlib}.}
\label{tab:artefacts}
\end{table}

The larger of two exclusions concerns 30 of the 210 responses, whose judge output did not
parse into ten clean verdicts. The released bank keeps these rows with blank judge fields,
and every statistic excludes them. Two questions lost all 14 of their responses and a
third lost two, and the scored bank spans 13 of the original 15 question clusters. The
release does not include the raw outputs behind the failures, leaving their cause among
truncation, formatting and refusal undetermined. Counting every unparsed verdict as an
error raises the per-element error rate from $4.72\%$ to $18.3\%$, an upper bound.
Treating an unparsed output as a rejection gives accuracies over all 210 responses of
$0.862$, $0.852$, $0.852$ and $0.881$ at cuts 4 to~7. Counting it as a gate error gives
$0.824$, $0.795$, $0.771$ and $0.786$, against $0.961$, $0.928$, $0.900$ and $0.917$ on the
180 scored responses. For the smaller exclusion, we dropped one model from an earlier
pilot after 17 of 40 calls returned a refusal stop reason. We did not reword or split the
prompt to work around a model-level content boundary.

\subsection{Bank construction and judge error structure}
\label{app:bank}

The bank behind Section~\ref{sec:setup} pairs every checklist element with a distractor,
and the pairing keeps most responses below the ceiling. Each of the 15 questions carries
its own fixed checklist of $K=10$ elements. Every element has a purpose-written
distractor that asserts something adjacent to the element or a common misconception on
the same sub-topic. We generate a response by drawing a random subset of elements to be
present. Each absent element's slot takes its distractor with probability $0.65$ and
otherwise stays empty. A judge that responds to length or topical coverage cannot separate
a distractor from the element it displaces.

The worked response in Table~\ref{tab:example} shows the judge rejecting a distractor and
accepting an element that the answer never states. The response includes five of the ten
elements for question~3, fills one absent slot with the distractor for element~9 and
leaves the other four empty. We identify this distractor by co-occurrence, as the released
bank does not label distractors. The sentence appears in 7 of the 14 responses to
question~3, and element~9 is absent from all seven. The judge rejects element~9 correctly
and returns nine correct verdicts in all. Its one error is to accept element~6, the
expansion valve, although the answer never mentions a valve and element~6's slot is empty.
The answer mentions expansion only through element~7.

\begin{table}[H]
\centering
{\small
\begin{tabular}{p{12.6cm}}
\toprule
\textbf{Question.} How does a refrigerator move heat out of its interior? \\
\midrule
\textbf{Answer as presented to the judge.} A refrigerant fluid circulates through a
closed loop of coils. Evaporation releases heat into the interior, which is then vented
outside separately. Expansion causes the refrigerant to cool to a temperature below the
interior. The cycle repeats continuously, net moving heat from cold interior to warm
exterior against the natural gradient. Heat release at the condenser causes the
refrigerant to condense into a liquid. The cold refrigerant absorbs heat from the
interior at the evaporator coils. \\
\bottomrule
\end{tabular}

\vspace{0.6ex}

\begin{tabular}{cp{10.6cm}cc}
\toprule
$j$ & Checklist element & $G_{ij}$ & $X_{ij}$ \\
\midrule
1 & A refrigerant fluid circulates through a closed loop of coils & 1 & 1 \\
2 & A compressor compresses the refrigerant vapor, raising its pressure and temperature & 0 & 0 \\
3 & Compression requires external work input, typically from an electric motor & 0 & 0 \\
4 & The hot compressed vapor releases heat to the room at the condenser coils & 0 & 0 \\
5 & Heat release at the condenser causes the refrigerant to condense into a liquid & 1 & 1 \\
6 & The liquid refrigerant passes through an expansion valve, dropping its pressure & 0 & \textbf{1} \\
7 & Expansion causes the refrigerant to cool to a temperature below the interior & 1 & 1 \\
8 & The cold refrigerant absorbs heat from the interior at the evaporator coils & 1 & 1 \\
9 & Heat absorption causes the refrigerant to evaporate back into a vapor & 0 & 0 \\
10 & The cycle repeats continuously, net moving heat from cold interior to warm exterior
against the natural gradient & 1 & 1 \\
\midrule
& Total & 5 & 6 \\
\bottomrule
\end{tabular}}
\caption{One response from the bank (\texttt{item\_id} 30, question~3), with gold
verdicts $G_{ij}$ and judge verdicts $X_{ij}$ for all ten elements, quoted verbatim from
the released file. An element with $G_{ij}=0$ is either absent or represented by its
distractor. Here the second sentence of the answer is the distractor for element~9. It
inverts the direction of heat flow, and the judge rejects element~9. The judge errs once,
accepting element~6, whose slot is empty.}
\label{tab:example}
\end{table}

\begin{figure}[H]
\centering
\includegraphics[width=\textwidth]{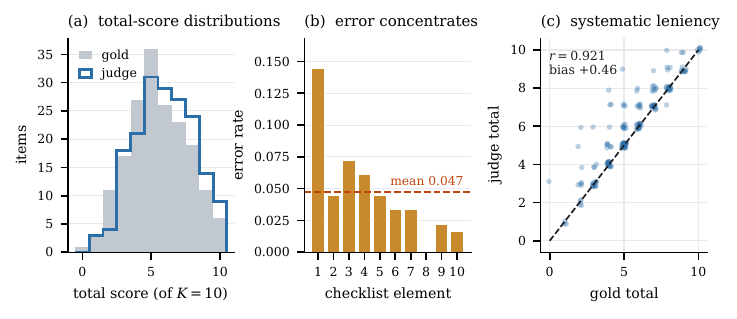}
\caption{Bank and judge diagnostics over the 180 scored responses. \textbf{(a)} Gold and
judge total-score distributions, with 6 gold totals and 9 judge totals at the ceiling of
10. \textbf{(b)} Judge error rate by element position, pooled over the 13 scored
questions, with the aggregate $4.72\%$ dashed. \textbf{(c)} Judge against gold totals with
the identity line, where the displacement above the line is the judge's leniency.}
\label{fig:bank}
\end{figure}

Across all 210 responses, gold totals have mean $5.45$ and SD $2.21$, with 2 responses at
0 and 6 at the ceiling of 10. On the 180 scored responses the mean is $5.51$ and the SD
$2.16$, and 6 gold totals and 9 judge totals reach the ceiling (Figure~\ref{fig:bank}a).
Table~\ref{tab:elements} gives realised presence rates and judge error by element position
over the scored responses.

\begin{table}[H]
\centering
{\small
\begin{tabular}{lcccccccccc}
\toprule
Element position & 1 & 2 & 3 & 4 & 5 & 6 & 7 & 8 & 9 & 10 \\
\midrule
Gold rate & 0.506 & 0.583 & 0.561 & 0.533 & 0.550 & 0.567 & 0.561 & 0.528 & 0.544 & 0.578 \\
Judge error & 0.144 & 0.044 & 0.072 & 0.061 & 0.044 & 0.033 & 0.033 & 0.000 & 0.022 & 0.017 \\
\bottomrule
\end{tabular}}
\caption{Gold presence rates and judge error rates by element position over the 180
scored responses. Each question has its own checklist, and each position pools 13
different lines. Presence rates stay close to balance by construction, whereas judge
error varies by an order of magnitude across positions.}
\label{tab:elements}
\end{table}

The judge's aggregate error of $4.72\%$ conceals three structures that matter for anyone
reusing the bank. First, the error runs almost entirely in one direction. The judge
accepts an absent element $10.4\%$ of the time ($95\%$ cluster interval
$5.5\%$--$15.5\%$) and rejects a present one $0.1\%$ of the time ($0.0\%$--$0.3\%$), and
the leniency of $+0.46$ elements consists almost entirely of accepted absent elements.
Second, the errors cluster within responses and questions. The variance of per-response error counts is $0.72$, against
$0.47$ if errors were independent at those rates, and three of the 13 questions hold 46 of the 85 errors. Independent errors at each checklist line's own rates would give a variance of $0.59$ and a mean KR-20 of $0.482$ (SD $0.019$). The judge's errors co-occur within responses beyond what line-level rates imply. Third, error concentrates on a few question-specific checklist lines. The 85 errors fall on 34 of the 130 lines, and the five worst lines hold 25 of them. Pooled
by position, the first-listed element errs at $0.144$ ($0.082$--$0.206$), with errors in 9
of the 13 questions. In practice, a gate calibrated on judge scores passes more responses
than the same gate calibrated on gold, and the five worst lines hold $29\%$ of the error.

A beta-binomial model with $\mathrm{Beta}(5.452, 3.663)$ describes the judge's total-score
distribution well, with $\chi^2 = 5.31$ on 6 degrees of freedom ($p = 0.504$). The fitted
model implies a reliability of $K/(\alpha+\beta+K) = 0.523$, the value that
Section~\ref{sec:misuse2} compares with KR-20. We report the fit to correct an earlier
draft's reading of overdispersion relative to a binomial as evidence of dependence between
elements. We withdrew that claim after noticing that the ratio of observed variance to
pooled binomial variance equals $1+(K-1)\rho$ by construction, with $\rho=1/(\alpha+\beta+1)$ the correlation between elements that the beta component implies. The ratio therefore
re-expresses the beta component's own parameter. The goodness of fit above tests only the
model's account of the total-score distribution. The model passes it, and the test cannot
detect dependence among elements.

\subsection{Estimators and procedures}
\label{app:estimators}

Every estimator operates on the element-level verdict matrix $X\in\{0,1\}^{n\times K}$ or
on its row totals $X_i=\sum_j X_{ij}$, and \texttt{estimators.py} implements each exactly
as stated here. Because each question has its own checklist, column $j$ of $X$ pools the 13
question-specific elements at position $j$. We treat the positions as the items of a
crossed design. With $\hat p_j=\frac{1}{n}\sum_i X_{ij}$ and $s^2_X$ the sample variance of
the totals,
\[
\mathrm{KR}\text{-}20(X)=\frac{K}{K-1}\Bigl(1-\frac{\sum_{j}\hat p_j(1-\hat p_j)}{s^2_X}\Bigr),
\]
and we compute it identically on $G$. Every term aggregates over responses and elements,
and no index runs over scorers. In that precise sense the coefficient carries no scorer
facet. The correlation $\phi = 0.063$ between element errors is the mean pairwise Pearson
correlation between the indicators $\mathbf{1}[X_{ij}\neq G_{ij}]$ across responses,
averaged over the pairs whose indicators vary, since element~8 has no errors. It inflates
the variance of per-response error counts by roughly $1+(K-1)\phi$. The same statistic on
the gold verdicts is $0.098$ and describes the bank.

Livingston--Lewis estimates fit a beta-binomial to the observed totals by maximum
likelihood, place a grid of 400 true proportions $p$ under the fitted
$\mathrm{Beta}(\alpha,\beta)$ density $w(p)$, and take
$\pi(p)=\Pr[X_i\geq c \mid p]=\sum_{x\geq c}\binom{K}{x}p^x(1-p)^{K-x}$ at cut $c$.
Classification accuracy and decision consistency are then
\begin{align*}
\mathrm{CA}&=\int w(p)\bigl[\pi(p)\,\mathbf{1}[pK\geq c]+(1-\pi(p))\,\mathbf{1}[pK<c]\bigr]\,dp,\\
\mathrm{DC}&=\int w(p)\bigl[\pi(p)^2+(1-\pi(p))^2\bigr]\,dp .
\end{align*}
Neither integral refers to $G$, and Section~\ref{sec:misuse2} rests on that indexing. This two-parameter form corresponds to the Livingston--Lewis procedure with true scores bounded by 0 and 1 and an effective test length of $K$, fitted here by maximum likelihood where Livingston and Lewis use moments. We apply the same function to the
gold totals for the error-free comparison in Table~\ref{tab:ll}.

The dependability index follows Brennan's estimator~\cite{brennan2001} for a design that
crosses responses with element positions. It targets
$\Phi(\lambda)=E_p(\mu_p-\lambda)^2/E_pE_I(X_{pI}-\lambda)^2$, the ratio of mean squared
distances from the pass mark for universe scores $\mu_p$ and observed proportions
$X_{pI}$. With variance components $\hat\sigma^2_p$,
$\hat\sigma^2_i$ and $\hat\sigma^2_{pi}$ from a two-way random-effects analysis of
variance,
\[
\Phi(\lambda)=\frac{\hat\sigma^2_p+(\bar X-\lambda)^2-\hat\sigma^2(\bar X)}
{\hat\sigma^2_p+(\bar X-\lambda)^2-\hat\sigma^2(\bar X)+\hat\sigma^2_\Delta},
\]
with $\lambda=c/K$ and $\bar X$ the grand mean proportion. The absolute error variance is
$\hat\sigma^2_\Delta=(\hat\sigma^2_i+\hat\sigma^2_{pi})/K$, and
$\hat\sigma^2(\bar X)=\hat\sigma^2_p/n+\hat\sigma^2_i/K+\hat\sigma^2_{pi}/(nK)$ corrects
the bias in $(\bar X-\lambda)^2$. The crossed treatment loses little on this bank, as gold
presence rates vary only from $0.506$ to $0.583$ across positions (Table~\ref{tab:elements})
and the estimated position component $\hat\sigma^2_i$ is below $0.0001$ on both sets of
verdicts. Variation among elements within a question enters $\hat\sigma^2_{pi}$, where
$\hat\sigma^2_\Delta$ counts it in full and KR-20 treats it as error on judge and gold
verdicts alike. Version 1 of this paper took the error term as the variance of all verdicts divided by $K$. That term includes response variance, and it understated $\Phi(\lambda)$ by $0.03$ to $0.05$ on the bank. Goodness of fit for the beta-binomial pools adjacent score
cells until every expected count reaches five and subtracts two fitted parameters from the
degrees of freedom.

Confidence intervals resample the 13 question clusters with replacement in $B=2000$ draws
and recompute the statistic on each draw. The independence checks apply element errors independently to the gold verdicts 2000 times, once at the overall false-accept and false-reject rates and once at each checklist line's own rates. The $\Phi(\lambda)$ control draws
$p_i\sim\mathrm{Beta}(5.452,3.663)$ and element verdicts $X_{ij}\sim\mathrm{Bernoulli}(p_i)$
for $n=4000$ responses. The Livingston--Lewis estimand control draws gold totals
$G_i\sim\mathrm{Binomial}(10,0.55)$ and a judge true score $\tau_i=0.9G_i+0.9+\varepsilon_i$
with $\varepsilon_i\sim\mathcal{N}(0,1.1^2)$, clipped to $[0,10]$. Observed totals are
$\tau_i+\eta_i$ with $\eta_i\sim\mathcal{N}(0,0.85^2)$, rounded to the nearest integer and
clipped to $[0,10]$. The control draws $n=4000$ responses and compares accuracy against
$\tau$ with accuracy against $G$ at each cut. The seeds are 11 for the two-way sweep, 13
for the measured-rate run, 17 for the one-directional sweep, 101 and 303 for the two
controls, 7 for the phrasing study and 0 for every bootstrap and for both independence
checks. These seeds reproduce the released script's output, and the published grids come
from an earlier stream that they match statistically (Appendix~\ref{app:repro}).

\subsection{Simulation grids and controls}
\label{app:grids}

A simulated bank at spread $s$ and judge error $e$ follows a three-step generative model.
Each response $i$ receives an element-presence rate
$\pi_i \sim \mathrm{U}(0.5 - s/2,\; 0.5 + s/2)$, gold verdicts follow
$G_{ij} \sim \mathrm{Bernoulli}(\pi_i)$, and the judge flips each verdict independently
with probability $e$. The spread $s$ thus controls between-response true-score variance
while leaving the mean at $K/2$. At $s=0$ the responses are exchangeable in difficulty, and
at $s=0.8$ the bank spans nearly the full score range. Table~\ref{tab:sweep} gives the grid
behind Figure~\ref{fig:sweep} and Section~\ref{sec:misuse1}, with 60 replicates per cell
at $n=210$ and $K=10$, and Table~\ref{tab:sweepsd} gives its per-cell standard deviations.

\begin{table}[H]
\centering
{\small
\begin{tabular}{lcccccc}
\toprule
& \multicolumn{6}{c}{Judge per-element error} \\
\cmidrule(l){2-7}
Bank spread & 0\% & 4.72\% & 5\% & 10\% & 20\% & 30\% \\
\midrule
0.0 & -0.000 & \textbf{0.007} & -0.014 & -0.011 & 0.015 & 0.003 \\
0.2 & 0.111 & \textbf{0.098} & 0.092 & 0.062 & 0.068 & -0.009 \\
0.4 & 0.366 & \textbf{0.317} & 0.328 & 0.253 & 0.174 & 0.045 \\
0.6 & 0.575 & \textbf{0.522} & 0.519 & 0.451 & 0.325 & 0.168 \\
0.8 & 0.722 & \textbf{0.679} & 0.682 & 0.608 & 0.457 & 0.266 \\
\bottomrule
\end{tabular}}
\caption{Mean KR-20 over 60 replicates per cell. Reading down any column shows bank
design moving the coefficient, and reading across any row shows judge error moving it.
The bold column comes from a separate 60-replicate run at the measured $4.72\%$ error rate
and is the column Section~\ref{sec:misuse1} quotes. The $5\%$ column belongs to the
two-way grid and comes from its own random stream, and the two columns differ by up to
$0.021$.}
\label{tab:sweep}
\end{table}

\begin{table}[H]
\centering
{\small
\begin{tabular}{lcccccc}
\toprule
Bank spread & 0\% & 4.72\% & 5\% & 10\% & 20\% & 30\% \\
\midrule
0.0 & 0.087 & 0.085 & 0.111 & 0.106 & 0.091 & 0.112 \\
0.2 & 0.097 & 0.083 & 0.097 & 0.078 & 0.093 & 0.100 \\
0.4 & 0.061 & 0.068 & 0.063 & 0.085 & 0.088 & 0.111 \\
0.6 & 0.044 & 0.047 & 0.048 & 0.041 & 0.065 & 0.084 \\
0.8 & 0.021 & 0.030 & 0.028 & 0.043 & 0.059 & 0.078 \\
\bottomrule
\end{tabular}}
\caption{Per-cell standard deviations for Table~\ref{tab:sweep}. Precision is lowest at
low bank spread, where the coefficient's true value lies near zero and its sampling
variance is largest. We leave the apparent non-monotonicity in the two lowest-spread rows
uninterpreted for that reason.}
\label{tab:sweepsd}
\end{table}

Two features of the grid support the identification claim more directly than the marginal
ranges in the main text. First, the grid holds cells with near-identical coefficients at
very different judge error rates. A spread of $0.6$ at $10\%$ error gives $0.451$, and a
spread of $0.8$ at $20\%$ error gives $0.457$, a difference of $0.006$ at twice the
error. Second, the effect of judge error depends on the bank, falling from $0.456$ at the
widest spread to nearly nothing at the narrowest. A practitioner therefore cannot calibrate
a correction without already knowing the bank's spread.

The measured judge errs almost entirely by accepting absent elements, whereas the grid's
judge flips verdicts symmetrically, and a one-directional sweep checks whether that
difference matters. At the measured rates of $10.4\%$ false acceptance and $0.1\%$ false
rejection, mean KR-20 over 60 replicates is $0.012$, $0.103$, $0.298$, $0.513$ and $0.671$
at spreads $0.0$ to $0.8$, close to the symmetric values of $0.007$ to $0.679$.
Lenient-only error at a per-element rate $e$ sets the false-accept rate to $2e$, since half
of all verdicts are absences at the mean presence rate of $0.5$. At the widest bank,
lenient-only error at rates of $0\%$, $5\%$, $10\%$, $20\%$ and $30\%$ gives $0.724$,
$0.674$, $0.630$, $0.495$ and $0.347$. The resulting span of $0.38$ compares with $0.46$
under symmetric error, and the conclusions of Section~\ref{sec:misuse1} stand.

The $\Phi(\lambda)$ control confirms that the gap between $\Phi(\lambda)$ and accuracy
reflects a mismatch of quantities. On data simulated from a beta-binomial with the fitted
parameters (Appendix~\ref{app:estimators}), Livingston--Lewis recovers classification
accuracy to within $0.02$ at every cut, while $\Phi(\lambda)$ departs from it by $0.14$ to
$0.25$. Estimation cannot explain the divergence, since the estimator hits its own target
in the same simulation where $\Phi(\lambda)$ diverges.

\begin{figure}[H]
\centering
\includegraphics[width=\textwidth]{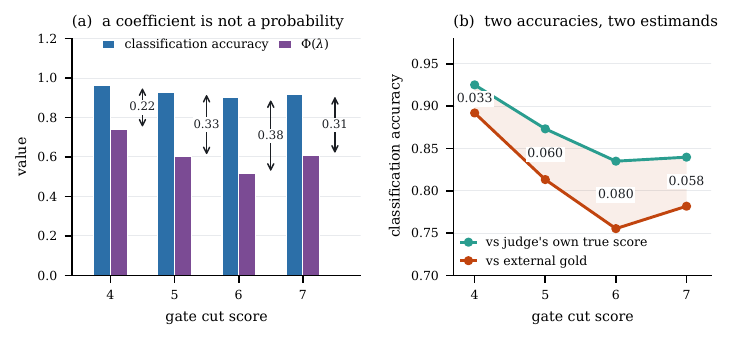}
\caption{Two quantities that resemble accuracies without being them. \textbf{(a)}
$\Phi(\lambda)$ against the judge's classification accuracy against gold at four cut
scores on the 180 scored responses, with arrows marking gaps of $0.22$ to $0.38$.
\textbf{(b)} In a simulation without element sampling, where the judge's true score
departs from gold, accuracy against the judge's own true score exceeds accuracy against
gold at every cut. On the real bank, element sampling reverses the sign of this difference
(Table~\ref{tab:ll}).}
\label{fig:estimands}
\end{figure}

Livingston--Lewis credits the error-free gold totals with accuracies of $0.75$ to $0.85$,
within $0.04$ of the judge's at every cut (Table~\ref{tab:ll}). Estimator error cannot
explain the difference column. The estimator recovers its own estimand to within $0.02$
in simulation, and the fitted model passes its goodness-of-fit test
(Appendix~\ref{app:bank}). The column equals the error-free shortfall less the judge's gate
error and less the difference between the two Livingston--Lewis estimates, and the
shortfall is the largest of the three terms at every cut. The column measures
mainly the distance between an estimand that treats the elements as a sample and a
criterion that holds them fixed.

\begin{table}[H]
\centering
{\small
\setlength{\tabcolsep}{4pt}
\begin{tabular}{cccccccc}
\toprule
& Accuracy & $95\%$ & LL CA & LL CA & Accuracy & $1-$LL CA & LL DC \\
Cut & vs gold & interval & judge & gold & $-$ LL CA & (gold) & judge \\
\midrule
4 & 0.961 & 0.922--0.994 & 0.882 & 0.846 & 0.079 & 0.154 & 0.836 \\
5 & 0.928 & 0.893--0.962 & 0.812 & 0.781 & 0.116 & 0.219 & 0.755 \\
6 & 0.900 & 0.851--0.945 & 0.759 & 0.750 & 0.141 & 0.250 & 0.697 \\
7 & 0.917 & 0.856--0.973 & 0.750 & 0.769 & 0.167 & 0.231 & 0.689 \\
\bottomrule
\end{tabular}}
\caption{Livingston--Lewis (LL) estimates on the 180 scored responses. Applied to the
error-free gold totals, the estimator falls short of perfect accuracy by more than the
judge's own difference at every cut. The bank's element sampling dominates the
difference column, and the judge's gate errors offset part of it.}
\label{tab:ll}
\end{table}

\subsection{The scorer facet: design and identification}
\label{app:phrasing}

A person-by-phrasing generalisability study can measure the scorer facet that the main text leaves unmeasured, and a simulation shows that the design recovers the phrasing component on average. Choi
et al.~\cite{choi2026} measure a related facet by giving each prompt variant its own item
parameters in a graded response model. Schroeder and Wood-Doughty~\cite{schroeder2024} take the repeated-administration route and compute McDonald's $\omega$ over repeated samples from one judge. Repeated scoring of the same responses by the same judge fails as a measure of the scorer facet for two reasons. On an earlier 20-response pilot with $K=8$ and two
administrations, the judge's scores did not change between runs, and the resulting
coefficient of $1.000$ would hold regardless of judge quality. Resampling at a fixed
prompt also measures decode-time sampling noise, a property of the sampler, and holds the
phrasing fixed. Reporting it as scorer reliability would repeat the estimand error of
Section~\ref{sec:misuse2}.

Deployments differ in rubric phrasing, since each commits to one wording of the scoring
instruction and inherits whatever variance that choice carries. Our design crosses $N=210$
responses with $R=3$ paraphrasings of the scoring instruction and holds checklist content
and responses fixed. The study estimates variance components from the three phrasings. We report $\Phi$ for absolute decisions at a single phrasing, as a
deployment commits to one and never averages over several. Like every reliability ratio,
$\Phi$ inherits the range restriction of Section~\ref{sec:misuse1} through
$\sigma^2(\text{responses})$, and a report should give the phrasing component beside it.
A simulation over 200 replicates confirms that the design recovers the components on
average in four scenarios, from irrelevant phrasing to a judge deterministic given its
phrasing (Table~\ref{tab:phrasing}). The inputs were
$\sigma^2(\text{phrasing})\in\{0,\,0.039,\,0.362\}$ with $\sigma^2(\text{resid})=0.090$
for the first three rows and $0.175$ with $\sigma^2(\text{resid})=0$ for the last, and the
table reports the recovered components.

\begin{table}[H]
\centering
{\small
\begin{tabular}{lcccc}
\toprule
Scenario & $\sigma^2(\text{responses})$ & $\sigma^2(\text{phrasing})$ & $\sigma^2(\text{resid})$ & $\Phi$ \\
\midrule
Phrasing irrelevant & 0.994 & 0.000 & 0.090 & 0.916 \\
Mild phrasing effect & 0.994 & 0.039 & 0.090 & 0.886 \\
Large phrasing effect & 0.994 & 0.365 & 0.090 & 0.721 \\
Deterministic given phrasing & 0.996 & 0.177 & 0.000 & 0.865 \\
\bottomrule
\end{tabular}}
\caption{Identification check over 200 simulations, $N = 210$, $R = 3$, by two-way
random-effects analysis of variance. $\Phi$ refers to a single phrasing and averages the
per-simulation values.}
\label{tab:phrasing}
\end{table}

The last row settles the case that ruled out the test--retest design. A judge perfectly
deterministic given a fixed phrasing has $\sigma^2(\text{resid}) = 0$, and the design
still identifies the phrasing component at $0.177$ with $\Phi = 0.865 < 1$ on average.
Three phrasings leave that component two degrees of freedom, and single studies scatter
widely. Across the 200 simulated studies its standard deviation is $0.183$, and $\Phi$
runs from $0.667$ to $0.988$ between the 5th and 95th percentiles. The phrasing study thus
yields an informative number in exactly the degenerate case where repeated administration
yields only $1.000$, although three phrasings leave that number imprecise.

\subsection{Worked reporting wording}
\label{app:template}

The four lines of Section~\ref{sec:gating} fit in a short disclosure, and
Table~\ref{tab:worked} fills them with this study's values for a practitioner to adapt.

\begin{table}[H]
\centering
{\small
\begin{tabular}{lp{10.4cm}}
\toprule
Line & Worked wording \\
\midrule
Bank & Reliability computed over $K=10$ binary checklist elements on 180 responses from 13
question clusters. SD of gold totals $2.16$ on a $0$--$10$ scale. Mean inter-element
correlation $0.098$. \\
\addlinespace
Gold comparison & KR-20 is $0.5223$ on judge verdicts and $0.5231$ on gold verdicts.
Livingston--Lewis accuracy at a pass mark of five elements is $0.812$ on judge totals and
$0.781$ on gold totals. Because error-free scoring lands in the same region, neither value
describes the judge. \\
\addlinespace
$\Phi(\lambda)$ & Dependability coefficient $\Phi(\lambda=0.5)=0.600$ on judge verdicts
and $0.543$ on gold verdicts. This is a ratio of squared distances from the pass mark and
is not comparable to the
classification accuracy of $0.928$ at the same pass mark. \\
\addlinespace
Judge error & Against gold, the judge accepts $10.4\%$ of absent elements ($95\%$ cluster
interval $5.5\%$--$15.5\%$) and rejects $0.1\%$ of present ones ($0.0\%$--$0.3\%$). Five
of the 130 checklist lines hold 25 of the 85 errors. Gate accuracy at a pass mark of five
elements is $0.928$ ($0.893$--$0.962$) on the 180 scored responses and $0.795$ on all 210
when unparsed outputs count as errors. \\
\bottomrule
\end{tabular}}
\caption{Worked wording for the four reporting lines of Section~\ref{sec:gating}, with
this study's values.}
\label{tab:worked}
\end{table}

\subsection{Glossary}
\label{app:glossary}

\begin{table}[H]
\centering
{\small
\begin{tabular}{lp{10.2cm}}
\toprule
Term & Meaning \\
\midrule
Response & One constructed short answer, the unit the judge scores; the released file
calls it an item \\
Bank & The set of responses a judge scores, here 210 constructed short answers \\
Element & One binary checklist criterion, with $K=10$ per question and a separate
checklist for each question \\
Element position & The index $j$ of an element within its question's checklist \\
Gold verdict $G_{ij}$ & Whether element $j$ is present in response $i$, fixed by
construction \\
Judge verdict $X_{ij}$ & The judge's YES ($1$) or NO ($0$) for element $j$ of response
$i$ \\
Facet & A source of variance that a measurement design allows to vary \\
Scorer facet & Variance that the scorer contributes, here the judge \\
Spread $s$ & Width of the uniform range of element-presence rates in a simulated bank \\
Cut $c$, pass mark $\lambda$ & Number of elements a response must satisfy to pass, and
$\lambda=c/K$ \\
KR-20 & Kuder--Richardson internal-consistency coefficient over the $K$ elements \\
$\Phi(\lambda)$ & Generalisability-theory dependability index for absolute decisions at
$\lambda$ \\
LL CA, LL DC & Livingston--Lewis classification accuracy and decision consistency \\
False-accept rate & Share of absent elements that the judge marks present \\
False-reject rate & Share of present elements that the judge marks absent \\
Leniency & Mean of judge total minus gold total, in elements \\
\bottomrule
\end{tabular}}
\caption{Terms and notation.}
\label{tab:glossary}
\end{table}

\end{document}